\documentclass[conference]{IEEEtran}

\makeatletter

\def\ps@IEEEtitlepagestyle{%
  \def\@oddfoot{\mycopyrightnotice}%
  \def\@evenfoot{}%
}
\def\mycopyrightnotice{%
  {\footnotesize 979-8-3315-7191-7/26/\$31.00~\copyright~2026 IEEE\hfill}
  \gdef\mycopyrightnotice{}
}

\usepackage{booktabs}
\usepackage{blindtext}
\usepackage{eso-pic}
\IEEEoverridecommandlockouts
\usepackage{cite}
\usepackage{amsmath,amssymb,amsfonts}
\usepackage{algorithmic}
\usepackage{graphicx}
\usepackage{textcomp}
\usepackage{multirow}
\usepackage{xcolor}
\def\BibTeX{{\rm B\kern-.05em{\sc i\kern-.025em b}\kern-.08em
    T\kern-.1667em\lower.7ex\hbox{E}\kern-.125emX}}
    
\usepackage{eso-pic}
\newcommand\AtPageUpperMyright[1]{\AtPageUpperLeft{%
 \put(\LenToUnit{0.17\paperwidth},\LenToUnit{-2cm}){%
     \parbox{0.9\textwidth}{\raggedleft\fontsize{8}{11}\selectfont #1}}%
 }}%
\newcommand{\conf}[1]{%
\AddToShipoutPictureBG*{%
\AtPageUpperMyright{#1}
}
}

\begin{document}
\title{\vspace*{1cm} Predicting Wave Reflection and Transmission in Heterogeneous Media via Fourier Operator-Based Transformer Modeling\\
}

\author{\IEEEauthorblockN{Zhe Bai}
\IEEEauthorblockA{\textit{AMCR Division} \\
\textit{Lawrence Berkeley National Lab}\\
Berkeley, USA \\
zhebai@lbl.gov
}
\and
\IEEEauthorblockN{Hans Johansen}
\IEEEauthorblockA{\textit{AMCR Division} \\
\textit{Lawrence Berkeley National Lab}\\
Berkeley, USA \\
hjohansen@lbl.gov}
}

\maketitle
\conf{\textit{  Proc. of International Conference on Artificial Intelligence, Computer, Data Sciences and Applications (ACDSA 2026) \\ 
5-7 February 2026, Boracay-Philippines}}
\begin{abstract}
We develop a machine learning (ML) surrogate model to approximate solutions to Maxwell's equations in one dimension, focusing on scenarios involving a material interface that reflects and transmits electro-magnetic waves.
Derived from high-fidelity Finite Volume (FV) simulations, our training data includes variations of the initial conditions, as well as variations in one material's speed of light, allowing for the model to learn a range of wave-material interaction behaviors.
The ML model autoregressively learns both the physical and frequency embeddings in a vision transformer-based framework.  
By incorporating Fourier transforms in the latent space, the wave number
spectra of the solutions aligns closely with the simulation data. Prediction errors exhibit an approximately linear growth over time with a sharp increase at the material interface. Test results show that the ML solution has adequate relative errors below $10\%$ in over $75$ time step rollouts, despite the presence of the discontinuity and unknown material properties.
\end{abstract}




\section{Introduction}

\subsection{Overview}
Electro-magnetic waves are ubiquitous in a wide range of scientific and engineering applications, including communications, photonics, material characterization and medical imaging. Accurately modeling how these waves propagate, reflect, and transmit across different media is critical for understanding and optimizing system performance. Maxwell's equations~\cite{Maxwell1892} are often the starting point when simulating 
    electro-magnetic (EM) devices, where combinations of complex geometries and 
    material changes are designed to produce a desired effect.
When simulating whole devices with nonlinear effects such as frequency-dependent materials or those with hysteresis, an inexpensive choice is the Finite Difference Time Domain method (FDTD)~\cite{Wu2009, Schneider2025}.
However, FDTD can have several drawbacks, including: dispersion errors, errors introduced near boundaries and material interfaces, and very long serial time integration, among others~\cite{Tidy3D2025,Schneider2025}.
These errors and bottlenecks then become issues in the ``inner loop'' of EM design optimization problems, for example, wave guide multiplexers~\cite{Mahlau2025,Lu2013} and nonlinear antennas~\cite{Meep2010}.
This work investigates a machine learning (ML) surrogate approach designed to predict the solution across \emph{all times} from only a few initial times steps.
The ML training data only includes the electric field, $E(x,t)$; the algorithm must infer the dynamics of $B(x,t)$ in Maxwell's equations from just the space- and time-evolution of one variable.
To evaluate the ML approach, we use training data from a 1D prototype of a transverse EM wave crossing a material boundary, and evaluate the accuracy of the ML solution.
Because 1D Maxwell's equations allows exact analytical solutions, we can assess the accuracy and spectral content at all times. 

\subsection{Related methods for Maxwell's equations}
Maxwell's equations can be simulated using FDTD methods, and also finite volume (FV)~\cite{Yee1997} and finite element (FE)~\cite{Couture2020} approaches in the time domain.
Spectral methods, including Fourier transforms-based approaches, can also be employed; however, nonlinear effects and material boundaries often limit their applicability except in special contexts.
Similarly, reducing Maxwell's equations to the wave equation can only be done in simplified circumstances.
FDTD methods typically use lower-order discretizations in space (e.g., second-order 
    differences on Yee-grids), and simple lower-order time integrators,
    such as the Leap Frog method \cite{Schneider2025}.
For FD and FV methods, material boundaries are often constrained to be on regular grid lines, 
    and implemented with lower-order or one-dimensional approximation~\cite{Meep2010}.
FE approaches can represent sharp material changes exactly by having a mesh aligned with
    material boundaries, but explicit time step constraints (CFL condition) can be 
    challenging with FE mesh refinement and higher-order discretizations.
As a result, FE methods typically lead to solvers in the frequency domain
    or use implicit time stepping~\cite{Couture2020}, which can be expensive or scale poorly.
Recent improvements in FV methods allow for sharp interfaces
    and higher-order accurate methods that do not have severe conditioning issues, which can otherwise impose restrictive time step constraints when material boundaries are 
    not aligned with the computational grid~\cite{OvertonKatz2023, Johansen2024}.
    
\subsection{Related ML methods for PDE solutions}
\label{ssec:relatedml}
Data-driven approaches provide a new paradigm to solve PDEs using simulation or experimental measurements that are amenable to leveraging physical constraints, such as physics-informed neural networks (PINNs)~\cite{raissi2019physics} that embeds governing laws into neural networks. However, PINNs may struggle in stiff loss landscapes with sharp gradients or discontinuities (e.g., material interfaces), and is limited to the geometry and boundary conditions that may not generalize to others. Neural operator methods~\cite{lu2021deeponet, li2023fourier, li2022transformer} build upon the universal approximation theorem, and effectively learn a mapping from the input functions to output functions, allowing them to learn PDE solutions with improved generalization.  
Recently, transformer-based models that leverage self-attention mechanism have shown more promise in modeling long-dependencies between input sequence elements and support parallel processing of sequence as compared to conventional types of recurrent networks~\cite{Khan2022transformer,Vaswani2017attention, Dosovitskiy2020ViT}.

In this work, we develop a Fourier operator-based surrogate model to emulate the evolution of electric field derived from Maxwell's equations using a dual-path neural architecture that incorporates both embeddings with self-attention from frequency and physical space, which enables the model to capture inherent dynamics for long-range predictions of wave propagation behaviors.

\section{Method}
\label{sec:method}
\subsection{Maxwell's equations and discretization}
\label{ssec:maxwellsfv}
Maxwell's equations govern the time evolution of electric field ($\mathbf{E}$) and magnetic 
    flux ($\mathbf{B}$), traveling at the speed of light in each material.
The simplest form normalizes against the speed of light in a vacuum, so that $c_m \le 1$ is a constant relative speed of light in each material $m$: 
\begin{align}
\label{eq:maxwell}
    \partial_t
    \begin{bmatrix}
        \mathbf{E} \\ 
        \mathbf{B}  
    \end{bmatrix}
    & = 
   \begin{bmatrix}
        {c^2_m} \nabla \times \mathbf{B} & 0 \\
        0 & -\nabla \times \mathbf{E}
    \end{bmatrix}  \, .
\end{align}
In the case of fields \emph{transverse} to the $x$ dimension this can be simplified to
    just one component, $\mathbf{E} = E\hat{y}$ and $\mathbf{B} = B\hat{z}$:
\begin{align}
\label{eq:maxwell1Dchar}
    \partial_t
    \begin{bmatrix}
        E \\ 
        B  
    \end{bmatrix}
    & = 
    -
    \begin{bmatrix}
        0 & {c^2_m} \\
        1 & 0
    \end{bmatrix}
    \partial_x 
    \begin{bmatrix}
        E \\ 
        B  
    \end{bmatrix} \\
\label{eq:maxwell1D}
    & =  -
    \partial_x 
    \begin{bmatrix}
        {c^2_m} B \\
        E
    \end{bmatrix}
\end{align}
    where this has been written in \emph{characteristic} \eqref{eq:maxwell1Dchar} and
    \emph{finite volume} \eqref{eq:maxwell1D} forms, respectively.
Using the \emph{method of characteristics} \cite{Evans1998} in 1D,
    the solution evolves along characteristics, 
    $x_0 = x \pm c_m t$, with 
    $E(x_0,t=0)$ and $B(x_0,t=0)$ contributing to the exact solution at all times.
This allows us to trace back from the current time, through or reflecting off 
    the material interface, to a value from the initial condition.
    
In finite volume form, we can use finite volume approximations that
    respect the transverse wave interface jump conditions, $[E] = 0$ and $[B] = 0$.
As is typical in FV methods, the $E$ and $B$ fields are transformed into left- and 
    right-moving characteristic variables; we use a fifth-order 
    upwind finite volume scheme that enforces jumps exactly at the interface.
A Riemann problem is solved on each finite volume cell interface,
    and at the material boundary the transmitted and reflected waves are calculated.
A flux-based update to $E$ and $B$ is used to evolve \eqref{eq:maxwell1D}
    with an  RK4 time integrator.
More details on this FVTD method can be found in~\cite{McCorquodale2015} where it is 
    applied to nonlinear compressible gas dynamics.

%
%
%
%

\subsection{Autoregressive modeling}
\label{ssec:pdeml}
\begin{figure}[htbp]
    \centering
\includegraphics[width=.95\columnwidth]{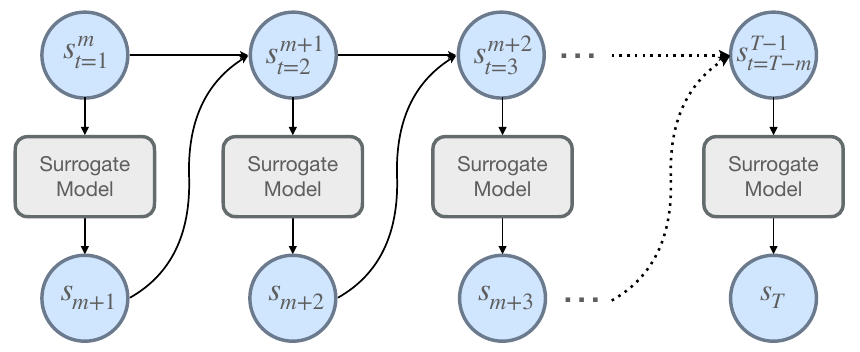}
    \caption{Autoregressive modeling for forecasting long-range dynamics with a series of time snapshots of $\tilde{s}_t = [s_t, s_{t-1}, \dots, s_{t-m+1}]$ at each step rolling out for prediction of $s_{t+1}$.}
    \label{fig:autoRM}
\end{figure}

To model the long-range evolution of the physical process, we consider an autoregressive formulation as illustrated in Fig.~\ref{fig:autoRM}. We define a vector $\tilde{s}_t$ that represents the previous snapshots of the variable at the timestamps from $t-m+1$ to $t$, in a total of $m$ ($m\ll T$) snapshots of $\tilde{s}_t = [s_t, s_{t-1}, \dots, s_{t-m+1}]$. The one-step forward neural operator is defined as
\begin{align}
        s_{t+1} = \mathcal{F}_{\theta_d}(\tilde{s}_{t}), \label{eq:onestep}
\end{align} 
where the operator $\mathcal{F}$ is composed of a dual-path Fourier Transformer that learns both the embeddings from the original physical space and the Fourier transformed space as in shown in Fig.~\ref{fig:model}.

\subsection{Fourier-based vision transformers}
\label{sec:fouriervision}
The model learns spatial scales through a hierarchy of Fourier transformer block that each composition block is designed to handle a specific scale. At each level, two transformation paths are performed: one is in the spatio-temporal domain, which operates self-attention on spatio-temporal embeddings; the other path is in the frequency domain, which operates on its transformed Fourier components.

\begin{itemize}

\item \emph{Frequency embeddings}: the model first applies a Fourier transform to the input tokens, retaining the low-frequency components. These components are then passed through a Fourier Transformed module to generate the frequency embeddings $f_{t}$. These embeddings are subsequently processed by a series of transformer layers that perform feature mixing and nonlinear transformations in the frequency domain. Finally, an inverse Fourier transform is applied, followed by a linear projection that maps the frequency-domain representations back to the spatial domain. 

\item \emph{Spatio-temporal embeddings}: the same set of tokens first pass through a spatio-temporal embedder, after which the spatio-temporal embeddings $e_{t}$ are processed by multiple standard transformer layers for mixing correlations and nonlinear transformation. Finally a separate linear projection is applied to get predictions for each patch of $s_{t+1}$.
\end{itemize}

\begin{figure}[htbp]
\centering
\includegraphics[width=\columnwidth]{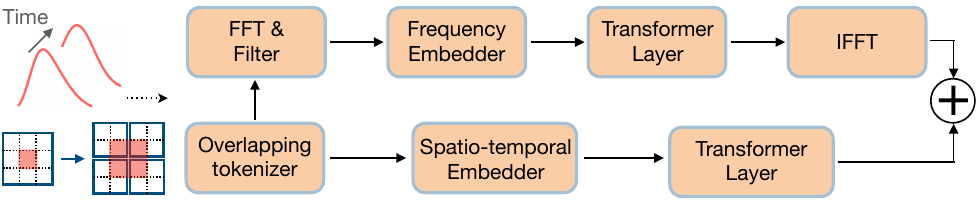}
\caption{Dual-path architecture: Fourier transformer. Overlapping tokenizer is applied prior to the transformations into embeddings of $f_{t}$ and $e_{t}$. After the two embeddings are merged, an overlapping detokenizer is employed to reconstruct the spatial field, mapping the latent representation back to the original physical domain.}
\label{fig:model}
\end{figure}

In addition, each Fourier transformer block first tokenizes the discretized functions $\tilde{s}_{t}$ along the spatial dimensions with an overlapping tokenizer (OLT), which allows adjacent patches to share boundaries through overlapping regions and enhance the spatial continuity. For instance, as shown in Fig.~\ref{fig:model}, a $3\times3$ input generates four $2\times2$ patches with a $1\times1$ overlap, where the overlapping areas (gray) are shared between patches. During detokenization, overlapping regions are reconstructed by averaging the corresponding values from neighboring patches in overlapping detokenizer (OLDT). This step mitigates discontinuity issues at the patch boundaries, improving the smoothness of continuous target functions incorporating shared boundaries into patch embeddings for sharp transitions.

\section{Results}
\label{sec:results}

\subsection{Maxwell's Equations Test Setup}
The domain $x \in [0,1]$ is split into two material regions, with speed of light
    $c_1 = 1$ for $x \le x_j = 0.5$, and $c_2$ for the right region, $x \ge x_j$.
The initial condition $\phi$ is entirely contained in the left half domain, 
    and is a right-moving wave packet with $E_0 = \phi$, $B_0 = \phi/c_1$.

\textbf{Test Case 1.}
We use two uniform random parameters $r_1, r_2 \in [0,1]$ 
    to generate $200$ test cases varying only the initial conditions:
\begin{align}
    \phi(x) & = 
    \begin{cases}
         g(x), & \hbox{ if } r_1 < 0.15 \\
         g(x) \sin(2 \pi k(x - x_s)), & \hbox{ otherwise.}
    \end{cases} \\
    \nonumber
   g(x) & = \exp(-(x - x_g)^2/\sigma^2), \hbox{ with } x_g = 0.25, \\
    \nonumber
    \sigma & = 0.25 / (5 - 2 r_1) \in [1/20, 1/12], \\
    \nonumber
    x_s & = 0.35 + 0.2 r_1 \in [0.35, 0.55], \\
    \nonumber
    k & = 3 + 3 r_2 \in [3, 6] \, .
\end{align}
For this test case, we fix $c_2 = 1 / 3$ in the second domain.
To generate accurate training data, we use the FV method described in \S\ref{ssec:maxwellsfv},
    with $N=256$ grid points, resulting in a pointwise
    relative error $< 0.1\%$ for all tests.
We run for $200$ time steps, and train the model across the time domain $[0,200]$.
 
\textbf{Test Case 2.}
Here we add a third uniform random variable, $r_3$, that specifies
    the speed of light in the second medium:
\begin{align}
    \label{eq:c2random}
    c_2 & = 1 / (1 + 2 r_3) \in [1/3,1]
\end{align}
    so that $c_2 \le c_1 = 1$.
This is an important variation to test if the ML approach can learn
    the speed of light directly from training data, and
    then \emph{infer} it from only the first few time steps.
Because $c_2$ has no influence before the wave packet has 
    crossed $x_j$, we first run for 100 time steps, then train the model on time steps 
    $[100,200]$, so that the effect of $c_2$ on the PDE solution can be observed as the wave packet crosses the material interface, $x_j$.
In both cases, the curated data is partitioned into $160$ samples for training, $20$ samples for validation and $20$ for test.

\subsection{Wave propagation prediction}
\begin{figure}[htbp]
    \centering    \includegraphics[width=.49\columnwidth, trim=0.75cm 0.1cm 0.cm 0.3cm,clip]{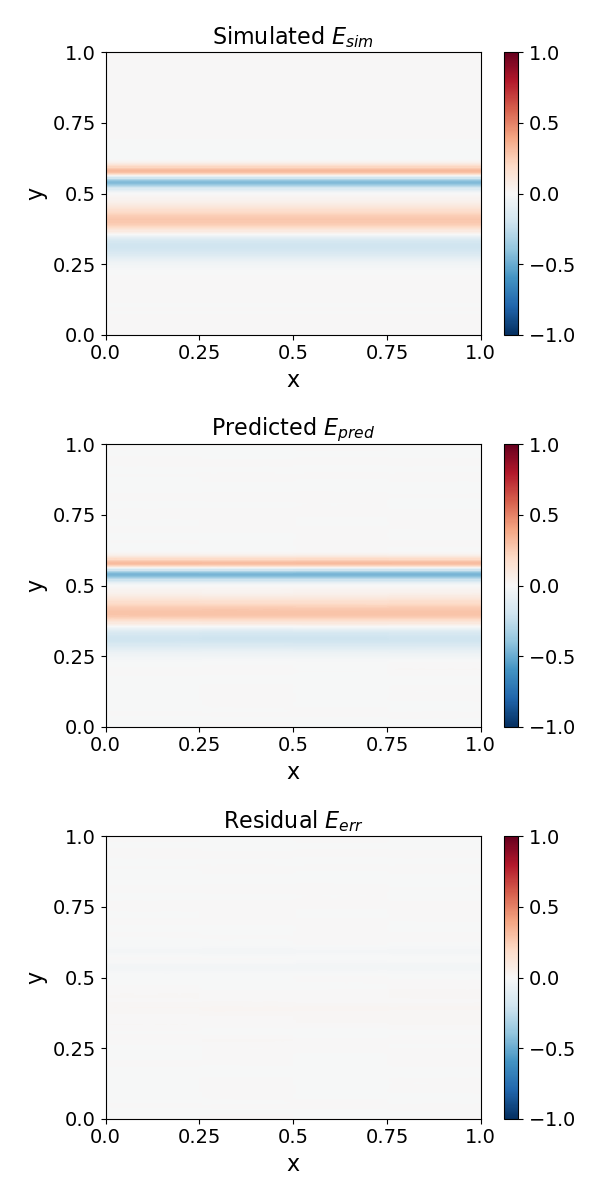}
\includegraphics[width=.49\columnwidth,trim=0.75cm 0.1cm 0.cm 0.3cm,clip]{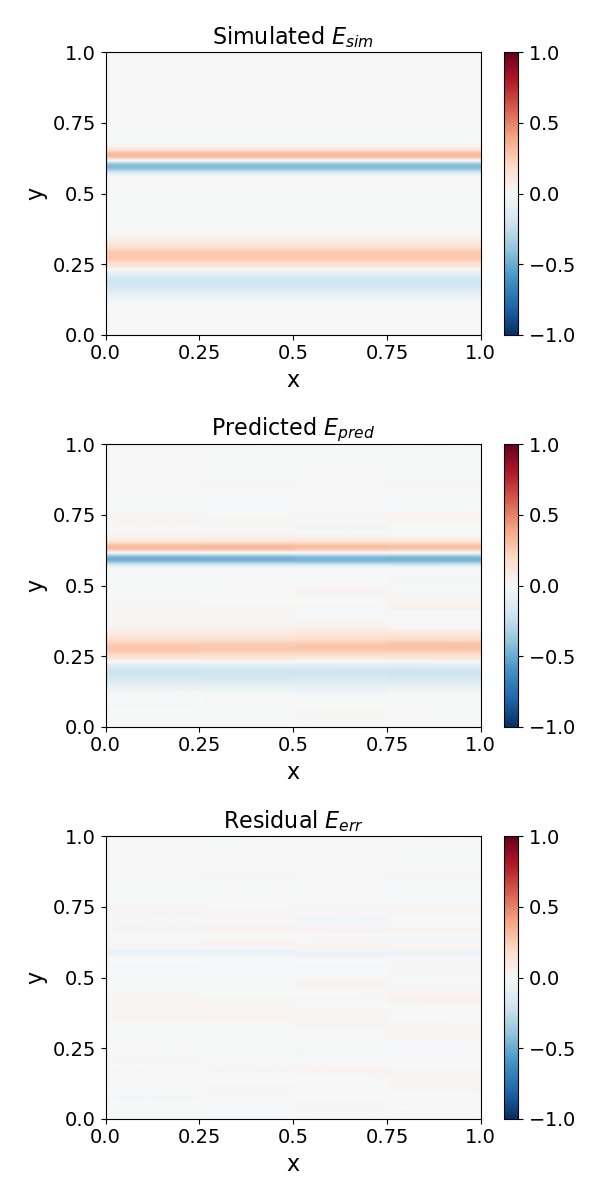}
\vspace{-.3in} 
\caption{Comparison between the simulated (top), ML predicted (middle), and residual (bottom) results for the electric field in Case 2, at $t=50$ (left column), and $t=100$ (right column). }
\label{fig:twoD}
\vspace{-.15in} 
\end{figure}

We investigate $E(x,t)$ over the entire time domain in 2D that is lifted from the 1D domain, in which values of one dimension are constant as exemplified in Fig.~\ref{fig:twoD}. 
Comparisons with the simulated results indicate that the ML model accurately captures the dynamics, with relative errors remaining small at earlier times ($t=50$) but increasing at later times ($t=100$), reflecting the temporal accumulation of approximation error.
 The ML model hyperparameters, including the network depth, hidden dimension, patch size are searched for optimized validation performance, as detailed in Table~\ref{tb:hyper_search}. Experiment results further indicate that incorporating the overlapping tokenizer alleviates sharp discontinuities at wave peaks introduced by parchification, leading to a notable improvement with relative error reductions of $20\%$. 
\vspace{-.09in}    
     
\begin{table}[htbp]
\centering
\caption{Hyperparameter search range}
\begin{tabular}{|c|c|}
\hline
\textbf{Hidden dimension} & $[256,512]$\\
\hline
\textbf{Patch size} & $[33,66,132]$\\
\hline
\textbf{Network depth} & $[6,10,12]$\\
\hline
\textbf{Overlap tokenizer} & $[0,1]$ \\
\hline
\end{tabular}
\label{tb:hyper_search}
\end{table}

\subsection{Error analysis - Test Case 1}
For Test Case 1, we fix the speed of light in the right half of the domain,
    $c_2 = 1/3$, and evaluate $200$ timesteps of the solution.
The initial wave packet begins in the left half of the domain, and crosses the material interface around step $t=100$.
By that point, the wave packet has started to reflect (invert) and travel left,
    as well as transmit into the right half of the domain, where it narrows
    substantially and continues traveling but at the slower speed $c_2$.
By $t=200$ both the reflected and transmitted waves have mostly moved away
    towards the boundaries.
 
\begin{figure}[htbp]
\centering
\includegraphics[width=.8\columnwidth,trim=0.75cm 0.65cm 0.5cm 0.5cm,clip]{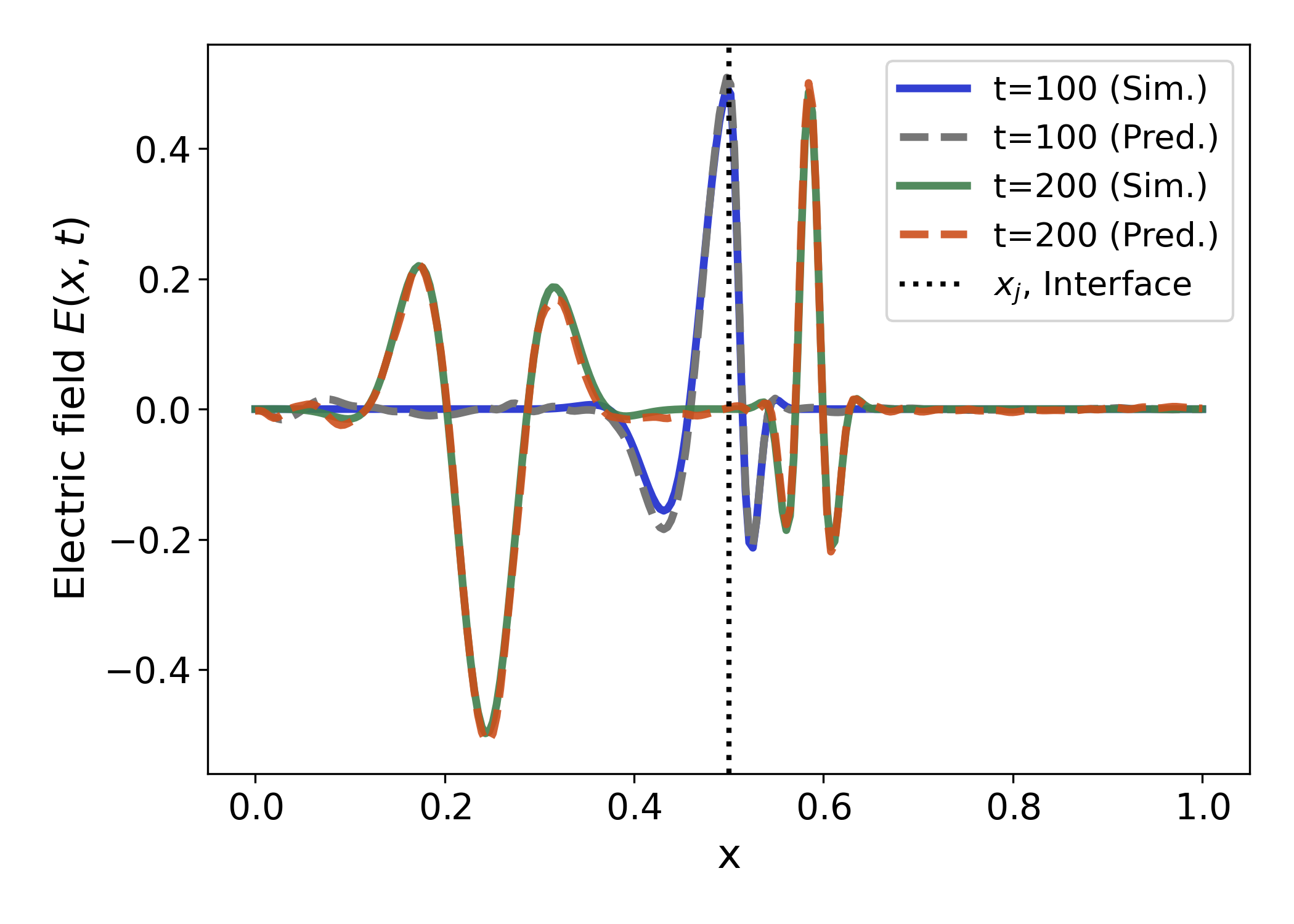}
\caption{Simulated and predicted solutions for Case 1, at two time steps
    $t=100$ and $t=200$. The dotted line at $x_j=0.5$ is the material
    interface where the speed of light changes.}
\label{fig:E_t-0}
\end{figure}
   
Fig.~\ref{fig:E_t-0} shows the simulation training data and the predicted ML solution at time steps $100$ and $200$.
Note that the predicted wave packets are of nearly the same profile and location,
    with only a small amount of visible amplitude error.
The final solution appears to match the transmitted and reflected waves as well.
There is also a small amount of higher wave-number error in the prediction throughout the domain.

\begin{figure}[hbtp]
    \centering
    \includegraphics[width=\columnwidth,trim=0.75cm 0.65cm 0.5cm 0.5cm,clip]{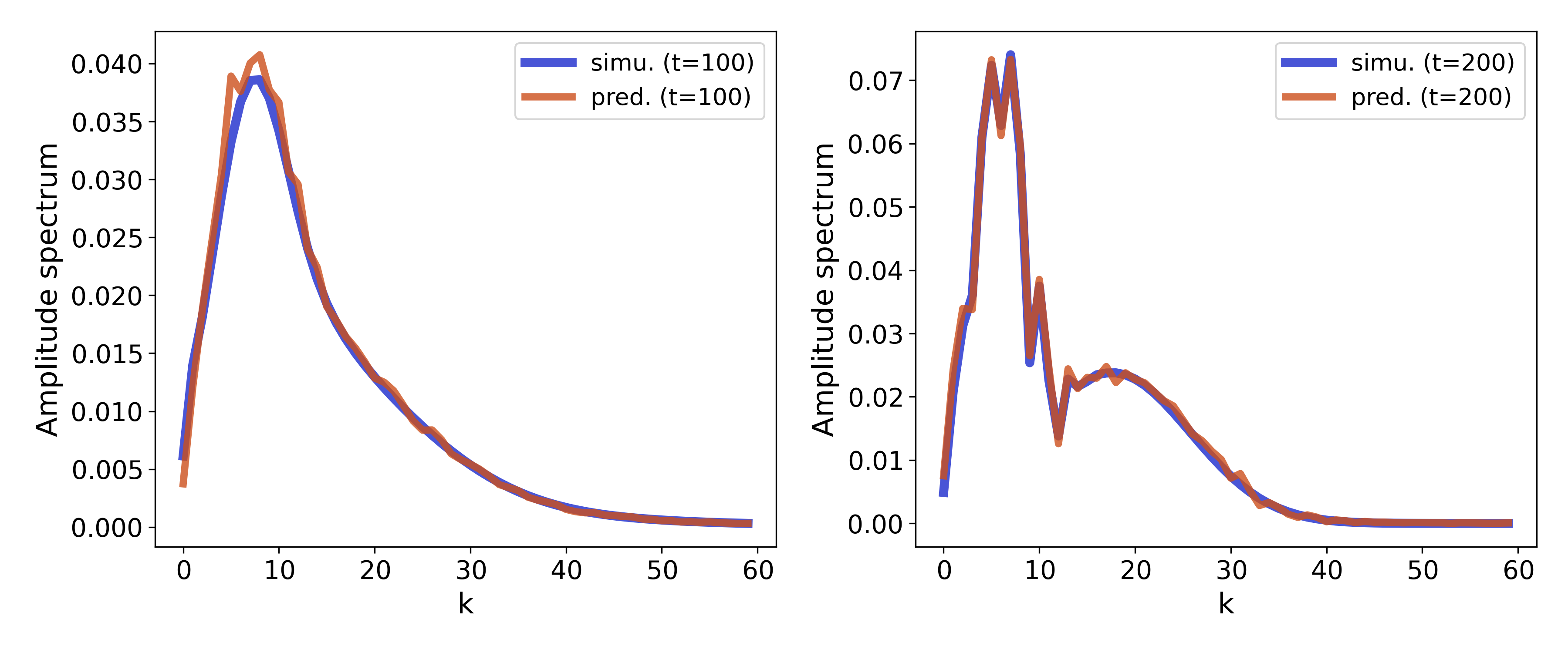}
    \caption{Fourier domain content of the predicted solution for Case 1,
    at step $t=100$ (left), when the wave packet is crossing the material boundary,
    and $t=200$ (right) when it has moved away after reflecting/transmitting.}
    \label{fig:fft-0}
\end{figure}

Fig.~\ref{fig:fft-0} shows the Fourier spectrum (wave mode content), 
    which supports these conclusions as well.
At timestep $100$, the predicted solution has largest relative errors at 
    the peak of the Fourier spectrum, but is overall consistent with
    the FVTD spectrum.
At timestep $200$, the spectrum matches better at the peak, even though it
    contains high wave number features from the narrower transmitted wave.
However, there is some noise throughout the spectrum in both cases, indicating
    that spatial errors are distributed relatively uniformly, everywhere.
    
\begin{figure}[htbp]
    \centering
\includegraphics[width=.85\columnwidth,trim=0.75cm 0.65cm 0.5cm 0.5cm,clip]{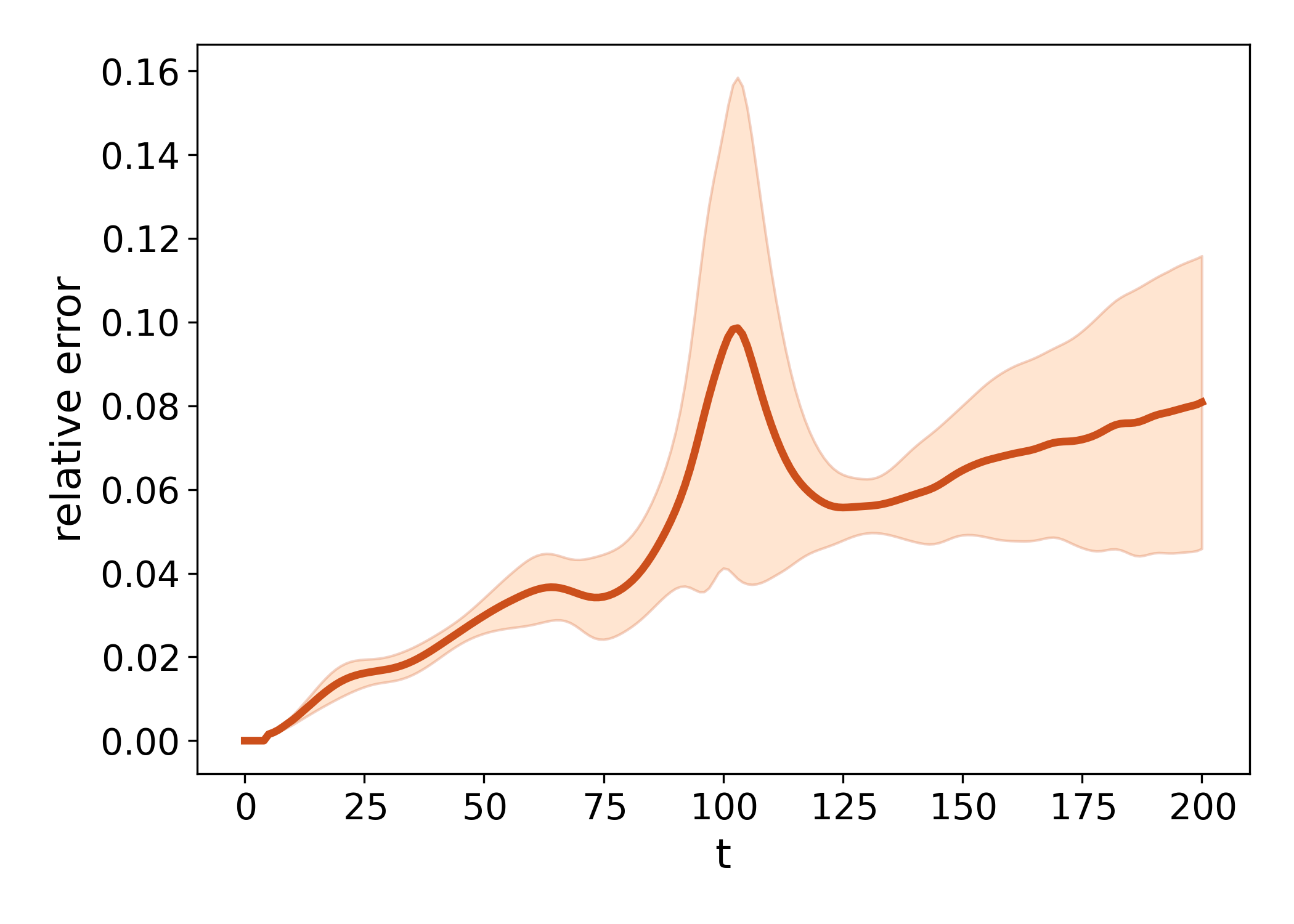}
    \caption{Case 1 predicted solution relative error over time steps $t=[0,200]$, 
        with the mean value (dark red line) and standard deviation (shaded
        region) across all test cases. Time step $t=100$ is approximately
        when the wave pulse is crossing the material boundary.}
    \label{fig:errbar-0}
\end{figure}

Fig.~\ref{fig:errbar-0} shows the relative error of the predicted solution
    over the whole time interval, $t=[0,200]$.
The mean appears to grow approximately linearly in time, except around step $t=100$,     
    where the error approximately doubles, and there is a much larger error distribution
    across the test data, reaching a relative error of around $\sim10\% \pm 5\%$.
This may not be surprising, as the wave packet develops a discontinuity as it
    moves through the material interface, which may affect its Fourier representation
    and require more modes than have been retained.
This is also the point where the reflected and transmitted wave packets
    are superimposed across the discontinuity, making it more difficult to separate
    the smooth spatial patterns that are clearer at earlier and later times.
 
For times after $t \approx 125$, the mean error reverts to the previous trend,
    but continues to grow, albeit with a much smaller variation across the test data.
In general, we see that the variation in error continues to grow more substantially over time,
    so that a significant portion of the tests are again around $\sim 10\%$
    relative error by the end of the time range.
Referring to Figs.~\ref{fig:E_t-0}-\ref{fig:fft-0} for comparison, the $E$ field amplitudes and spectral content may still match relatively well across the time domain.
 
\subsection{Error analysis - Test Case 2}
For this test, the speed of light $c_2$ in the right half of the domain varies randomly
    across individual test cases.
This affects both the amplitude of the transmitted and reflected waves, as well as
    the speed that the transmitted wave propagates.
For the algorithm to reliably infer $c_2$ from the time-series data, it must observe the dynamics over several time steps. Accordingly, we start the training data at time step $t=100$,
which corresponds to the onset of observable transmitted and reflected waves.
The time variable is then reindexed starting at this point $t=0$, and continued for an additional $100$ time steps subsequently.

\begin{figure}[htbp]
\centering\includegraphics[width=.8\columnwidth,trim=0.75cm 0.65cm 0.5cm 0.5cm,clip]{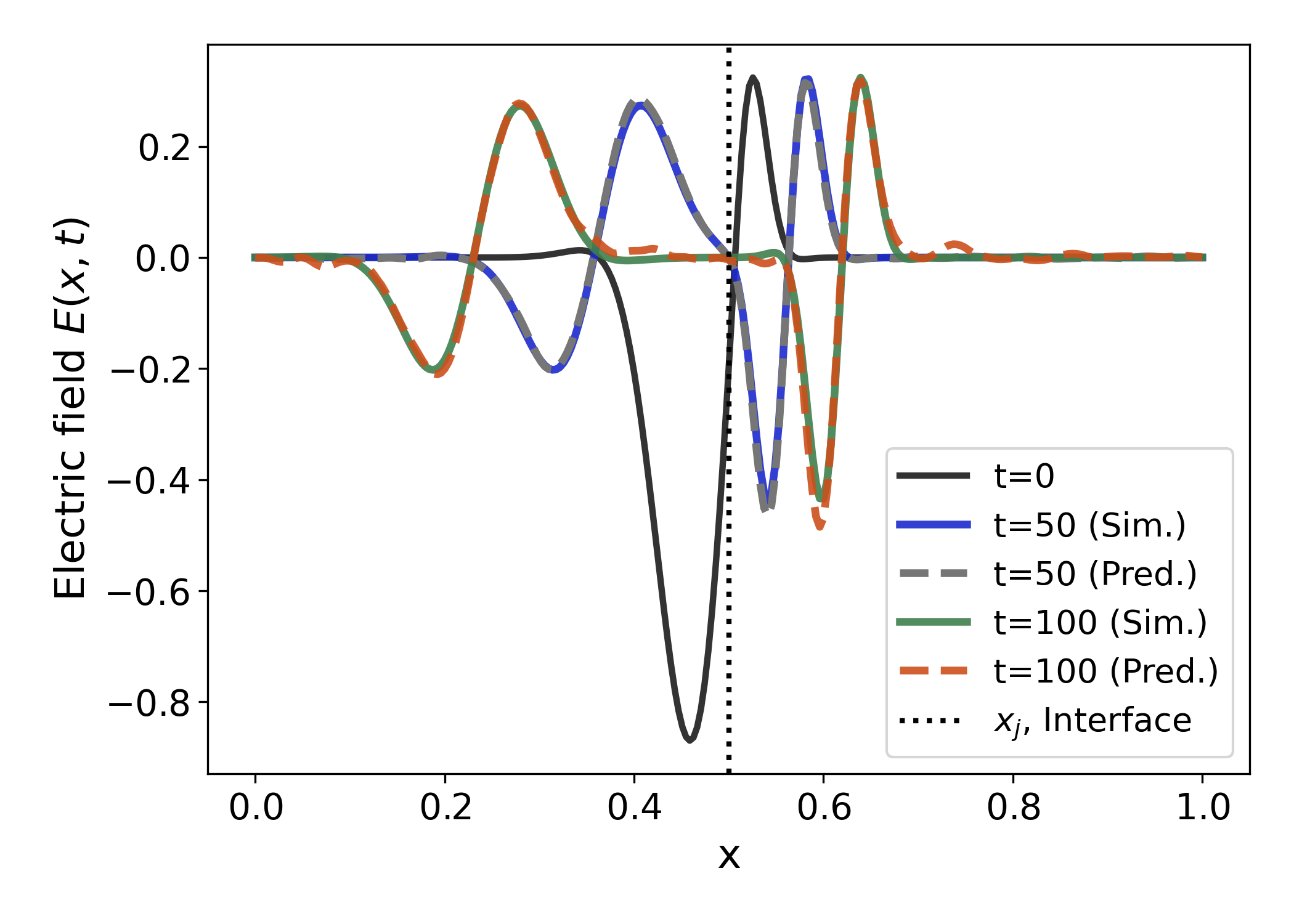}
    \vspace{-.1in}
    \caption{Simulated and predicted solutions for Case 2, with the initial condition and two time steps
    $t=50$ and $t=100$. The material to the right of the interface (dotted line) has a random speed of light that must be learned from training data.}
    \label{fig:E_t-1}
\end{figure}

\begin{figure}[htbp]
\centering
\includegraphics[width=\columnwidth,trim=0.75cm 0.65cm 0.5cm 0.5cm,clip]{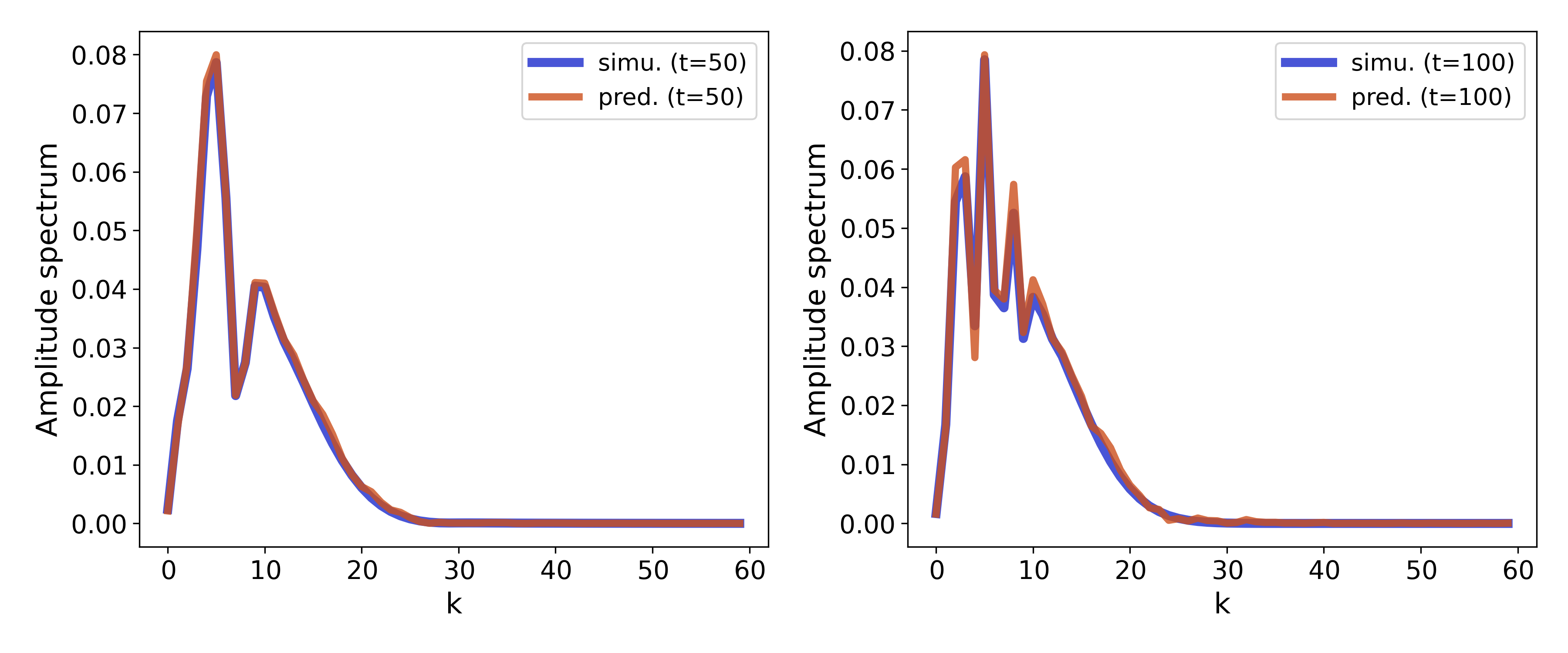}
    \vspace{-.2in} 
    \caption{Fourier domain content of the predicted solution for Case 2,
    at step $t=50$ (left) and $t=100$ (right), as it is moving away
    from the interface after reflecting/transmitting.}
    \label{fig:fft-1}
\end{figure}

Referring to Fig.~\ref{fig:twoD}, we see a color plot of the 2D solution
    at the initial condition (time-shifted, $t=0$) and later steps $t=50$ and $t=100$.
The wave packet location and amplitude are well-predicted,
    and the error at later times is larger, as before.
As illustrated in Fig.~\ref{fig:E_t-1}, a similar trend is observed: the ML model accurately reproduces the phase evolution of the wave packet, but the peak amplitude is less precise.
Note that the initial condition already contains some structure in the right half of the domain,
    and because the first few time steps are used for inference, the model is able to determine an
    approximate $c_2$.
However, the spectral content is not as clearly discernible from the initial condition due to the superimposition of
incident and reflected waves in the left domain.
As a result, the model must infer the propagation during training, yet it predicts the transmitted and reflected waves relatively well at later times, for example at $t=100$.
The Fourier content, seen in Fig.~\ref{fig:fft-1}, matches well in both wave number
    and amplitude at both times, as previously observed in Case 1.
    
\begin{figure}[htbp]
    \centering
    \includegraphics[width=.85\columnwidth,trim=0.75cm 0.65cm 0.5cm 0.5cm,clip]{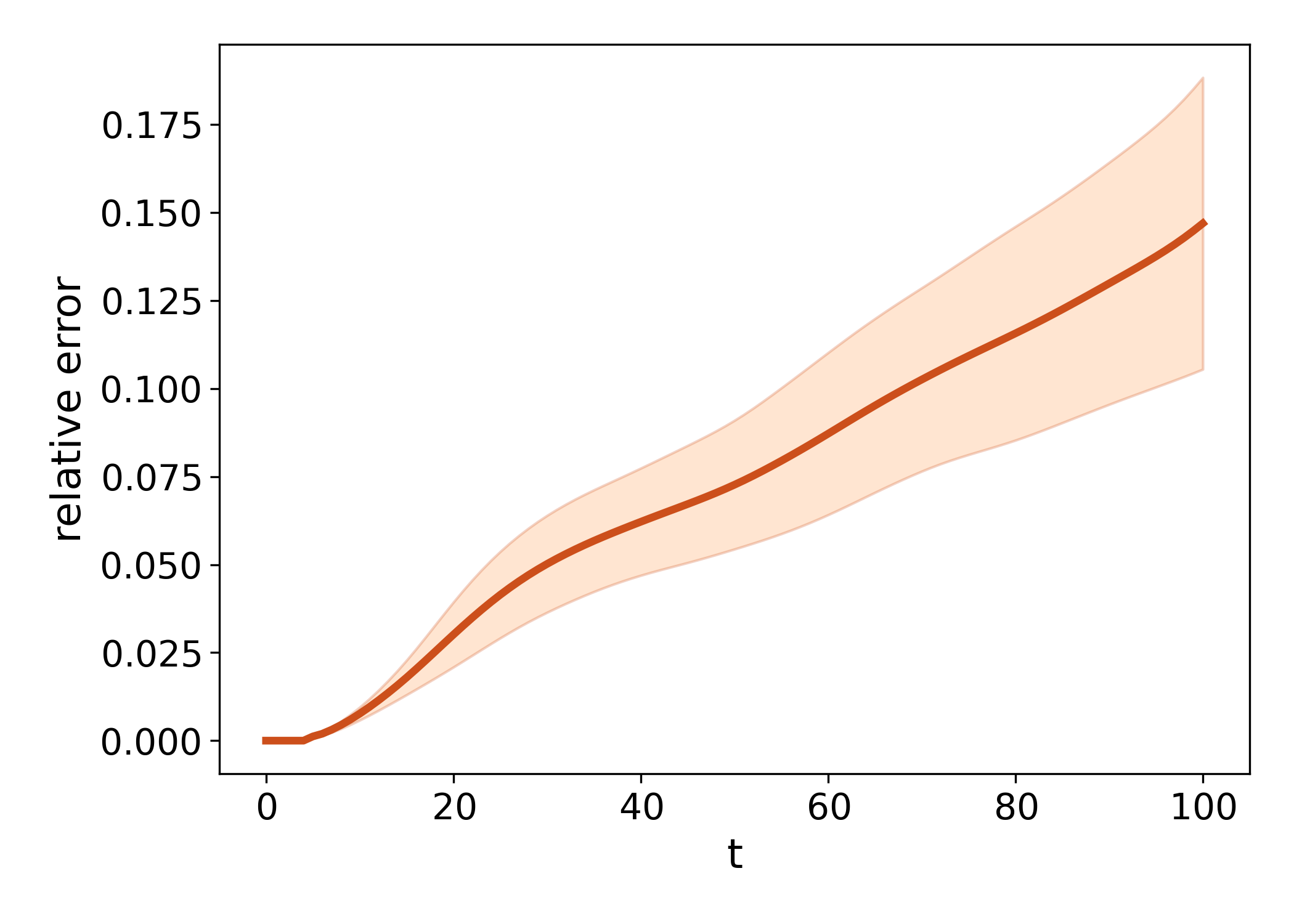}
    \vspace{-.1in} 
    \caption{Case 2 predicted solution relative error for $t\in[0,100]$, 
        mean value (dark red line) and standard deviation (shaded region) across all test data. The error grows approximately linearly in time.}
    \label{fig:errbar-1}
\end{figure}

A more substantial difference from Case 1 is seen in the time evolution 
    of the relative error, in Fig.~\ref{fig:errbar-1}.
The error and its variation in time are more consistent,
    but grow at approximately $2 \times$ the rate of Case 1, 
    effectively $4 \times$ higher with half the number of steps.
We suspect it is more challenging to start the training data when the wave pulse
    is discontinuous crossing the interface, especially with unknown material ($c_2$ parameterized) in the second medium.
So although the randomly varying speeds of light are immediately 
    evident to the model, it does exhibit greater errors overall.
Still, the short-term errors are small and the error growth
    is more steady than in Case 1.

\vspace{-.1in}
\subsection{Ablation study}
\subsubsection{The Frequency Path}
To evaluate the effectiveness of the frequency path embeddings, we conduct an ablation study removing the frequency path in each hierarchical layer. The experimental results are summarized in Table~\ref{tb:ablation0}, with all experiments trained for the same number of epochs. The $L_2$ relative error drops to $0.0506$ from $0.3945$ in Case 1, and for Case 2 with a half test time window, the relative error drops from $0.1819$ to $0.0827$.

\vspace{-.1in}
\begin{table}[htbp]
\centering
\caption{Effect of the frequency path.}
\begin{tabular}{c|c|c}
\toprule
\textbf{Problem} & \textbf{Model(s)} & \textbf{$L_2$ error}\\
\hline
\multirow{2}{*}{Case 1}
& ${f}_t + {e}_t$  & \textbf{0.0506}
 \\
& ${e_t}$ & 0.3945 \\
\hline
\multirow{2}{*}{Case 2}
& ${f_t} + {e_t}$  & \textbf{0.0827}
 \\
& ${e_t}$ & 0.1819 \\
\bottomrule
\end{tabular}
\label{tb:ablation0}
\end{table}
\vspace{-.1in}
\begin{table}[htbp]
\centering
\vspace{-.1in}
\caption{Ablation results on the Fourier mode truncation.}
\begin{tabular}{c|c|c}
\toprule
\textbf{Problem} & \textbf{Setting(s)} & \textbf{$L_2$ error}\\
\hline
\multirow{3}{*}{Case 1}
& $r=4$  & 0.0825
 \\
& $r=8$ &  \textbf{0.0506}\\
& $r=16$ &  0.4795\\
\hline
\multirow{3}{*}{Case 2}
& $r=4$  & 0.1191
 \\
& $r=8$ &  \textbf{0.0827}\\
& $r=16$ &  0.3099\\
\bottomrule
\end{tabular}
\label{tb:ablation0}
\end{table}
\subsubsection{The Fourier modes}
We also investigate the number of truncated Fourier modes in the Frequency path $f_t$. It is observed that $r=8$ yields the best overall performance. When fewer modes are retained, the model fails to adequately capture the dominant peak frequencies (see Figs.~\ref{fig:fft-0} and~\ref{fig:fft-1}). In contrast, retaining a larger number of frequency modes does not further improve accuracy, as the expanded frequency range exhibits reduced effectiveness in resolving the fine-scale structures compared to the spatio-temporal path.

\section{Conclusion}
We have presented a Fourier operator-based Transformer method as a surrogate model for predicting the 
    time series solution to 1D Maxwell's equations with a material boundary.
With the results of two test cases, it is shown that the ML approach is able
    to predict solution to a relative error of $\sim10\%$ 
    across most of the time domain.
The errors are much smaller early in the simulation, and grow approximately
    linearly through later time steps, with a growth in variation of errors across
    the test data as well.
More importantly, the Fourier domain representation of the solution is also
    accurate, representing the dominant wave numbers well and maintaining the solution's form throughout.
    
The ablation study results demonstrate the effectiveness of the frequency path in capturing the dominant spectral components of the system. In particular, the optimal number of retained modes in the frequency path corresponds closely to the peak frequencies in the spectrum. 
For the first test case, where the wave packet crosses the material boundary
    half way through the simulation, the ML prediction error increases briefly,
    then decreases again in later time steps.
We suspect that this is due to the material discontinuity introducing high wave-number
    content that is not in the truncated Fourier representation.
The second test case demonstrates that, if the model is trained on data that includes
    solution values in both material domains, it greatly reduces this error at
    the interface.
The model is then able to recover long-term accuracy, even when the material property is
    random distributed within the test data, implying that it effectively infers the hidden variable or characteristics of the media.
    
In future work, building on our understanding of the limitations and sources of error in the present approach,
    we will extend it to higher dimensions and nonlinear materials, as seen in waveguide multiplexers \cite{Lu2013}.
An additional consideration is the assessment of computational gains offered by the ML model over traditional low-accuracy finite-volume models. From a scaling standpoint, the ability to predict the entire
    time evolution at once would vastly improve parallelism over traditional sequential time stepping;
    this could open the door for doing parallel in time algorithms \cite{Ibrahim2025Parareal}, as well.
Finally, we expect that these time series predictions may lead to better initial guesses
    in solver preconditioners and inverse design problems of these complex systems.

\section*{Acknowledgment}
 This material is based upon work supported by the U.S. Department of Energy, Office of Science, Office of Advanced Scientific Computing Research under contract number DE-AC02-05CH11231. 
 Z.B. acknowledges support from the U.S. Department of Energy, Office of Science, SciDAC/Advanced Scientific Computing Research under Award Number DE-AC02-05CH11231. 
 This research used resources of the National Energy Research Scientific Computing Center (NERSC), a Department of Energy User Facility.

\end{document}